%% file: elsarticle-template-num-names.tex
\documentclass[preprint,5p,twocolumn]{elsarticle}

\usepackage{booktabs,ragged2e}
\usepackage[flushleft]{threeparttable}
\usepackage{multirow}
\usepackage{amsmath,amsfonts}
\usepackage{algorithmic}
\usepackage{algorithm}
\usepackage{array}
\usepackage[caption=false,font=normalsize,labelfont=sf,textfont=sf]{subfig}
\usepackage{textcomp}
\usepackage{stfloats}
\usepackage{url}
\usepackage{verbatim}
\usepackage{graphicx}
\usepackage{xcolor}
\usepackage{hyperref}
\hypersetup{
	pdfborder = {0 0 0},
	pdfauthor = {Mario Villaizán-Vallelado, Matteo Salvatori, Belén Carro Martinez, Antonio Javier Sanchez Esguevillas},
	pdftitle={Graph Neural Network contextual embedding for Deep Learning on Tabular Data}
}
\usepackage[acronym]{glossaries}
\input{glossaries}

\usepackage{amssymb}
\journal{Knowledge-Based Systems}

\begin{document}

\begin{frontmatter}



\title{\acrlong{gnn} Contextual Embedding for \acrlong{dl} on Tabular Data}


\author[ailab,uva]{Mario Villaizán-Vallelado}
\ead{mario.villaizanvallelado@telefonica.com}
\ead{mario.villaizan@uva.es}

\author[ailab]{Matteo Salvatori}
\ead{matteo.salvatori@telefonica.com}

\author[uva]{Belén Carro Martinez}
\ead{belcar@tel.uva.es}

\author[uva]{Antonio Javier Sanchez Esguevillas}
\ead{antoniojavier.sanchez@uva.es}

\affiliation[ailab]{
	organization={\acrlong{ai} Laboratory (\acrshort{ai}-Lab), Telefonica I+D},
	country={Spain}}

\affiliation[uva]{
	organization={Universidad de Valladolid},
	city={Valladolid},
	postcode={47011},
	country={Spain}}

\begin{abstract}
All industries are trying to leverage \acrfull{ai} based on their existing big data which is available in so called tabular form, where each record is composed of a number of heterogeneous continuous and categorical columns also known as features. \acrfull{dl} has constituted a major breakthrough for \acrshort{ai} in fields related to human skills like natural language processing, but its applicability to tabular data has been more challenging. More classical \acrfull{ml} models like tree-based ensemble ones usually perform better. This paper presents a novel \acrshort{dl} model using \acrfull{gnn} more specifically \acrfull{in}, for contextual embedding and modelling interactions among tabular features. Its results outperform those of a recently published survey with \acrshort{dl} benchmark based on five public datasets, also achieving competitive results when compared to boosted-tree solutions.
\end{abstract}




\end{frontmatter}


\section{Introduction}

Many practical real-world applications store data in tabular form, i.e. samples (rows) with the same set of attributes (columns). Medicine, finance or recommender systems are some common examples.  

\acrshort{dl} success in tasks involving texts, images or audio has sparked interest in its possible application to tabular data. Nevertheless, this success is often achieved when the  input data are homogeneous and the structure used to organize the information provides insights about the data understanding.  All tokens in a sentence are instances of the same categorical variable and their layout has semantic significance. Pixels in an image are continuous and usually have spatial correlation. 

Tabular data have two characteristics that hinder \acrshort{dl} performance. On one hand, tabular features are heterogeneous, having a mix of continuous and categorical distributions that may correlate or be independent. On the other hand, the meaningfulness of tabular data row is independent of the column order, i.e. position is arbitrary and does not provide information.

Tree-based ensemble models such as \acrshort{xgboost} \cite{chen2016xgboost}, \acrshort{catboost} \cite{prokhorenkova2018catboost}, and \acrshort{lightgbm} \cite{ke2017lightgbm} achieve the \acrfull{sota} performances on tabular data: they have competitive prediction accuracy and are fast to train. However, further research and development of \acrshort{dl} models for tabular data are motivated, by the fact that standard tree-based approaches have limitations, for example, in case of continual learning, reinforcement learning or when tabular data is only part of the model input, which also includes data such as images, texts or audio.  

Inspired by the success of contextual embedding in large language models (for example \acrshort{bert} \cite{devlin2019bert}), several recent research \cite{huang2020tabtransformer, gorishniy2021revisiting, somepalli2021saint} have investigated how to enhance tabular feature representation (and hence global \acrshort{dl} model performances) by taking into consideration their context, that is, feature interaction. The results obtained in these works, as well as the outcomes of recent comparisons on many public datasets \cite{borisov2022deep}, illustrate how the contextual embedding approach tends to outperform not only standard \acrfull{mlp} models, but also more complex models developed to solve complicated tasks \cite{he2017neural, guo2017deepfm, cheng2016wide, naumov2019deep, wang2021dcnv2} or models combining \acrshort{dl} architectures with standard \acrshort{ml} approaches \cite{popov2019neural, arik2021tabnet}. 

Many of the most recent studies employ Transformers \cite{vaswani2017attention} as a method for contextual embedding. However, in this paper, we look at how to use a \acrshort{gnn} to improve contextual embedding for tabular data. \acrshort{gnn}s are a special subset of neural networks that are capable of managing information organized in a graph which is a structure with variable shape or size and with complex topological relations. One of the most important features of a graph is that its meaning does not depend on the order of its nodes, just as the meaning of a tabular row does not depend on the order of its columns. 

\textbf{Contributions}. The contributions of our paper are summarized as follows: 
\begin{itemize}
	\item  We introduce \acrfull{ince}, a \acrshort{dl} model for tabular data that employs \acrshort{gnn}s and, more specifically, \acrlong{in}s \cite{battaglia2016interaction, battaglia2018relational, sanchezgonzalez2020learning} for contextual embedding. First, all features (categorical and continuous) are individually projected in a common dense latent space. The resultant feature embedding is organized in a fully-connected graph with an extra virtual node, called \acrshort{cls} as in \acrshort{bert} \cite{devlin2019bert}. Then, a stack of \acrshort{in}s models the relationship among all the nodes - original features and  \acrshort{cls} virtual node - and enhances their representation. The resulting \acrshort{cls} virtual node is sent into the final classifier/regressor. For sake of reproducibility, we share an implementation of INCE\footnote{\href{https://github.com/MatteoSalvatori/INCE}{https://github.com/MatteoSalvatori/INCE}}\footnote{\href{https://codeocean.com/capsule/2256574/}{https://codeocean.com/capsule/2256574}}.
	\item We compare \acrshort{ince} against a wide range of deep tabular models and generally used tree-based approaches, using the tabular datasets provided in \cite{borisov2022deep} as a benchmark. \acrshort{ince} outperforms all other \acrshort{dl} methods on average, and it achieves competitive results when compared to boosted-tree solutions.
	\item We thoroughly investigate the differences between contextual embeddings based on Transformers and \acrshort{in}s and analyze the influence of \acrshort{in} hyperparameters on model performance: quality of results, model size, computational time. Regardless of the dataset or task challenge, we gain a collection of patterns that aid in the establishment of a strong baseline.
	\item We investigate the interpretability of the \acrshort{in} ensuing contextual embeddings. On the one hand, we focus on the feature-feature relationship discovered by the \acrshort{in},  while on the other hand, we concentrate on how contextual embeddings improve traditional context-free embeddings.
\end{itemize}

\section{Related Work}
\label{sec:related_work}

\textbf{Standard Tabular Models}. As already commented, when dealing with tabular data, tree-based ensemble models such as \acrshort{xgboost}, \acrshort{catboost} and \acrshort{lightgbm} are often a popular choice. They usually provide high performance regardless of the amount of data available, as they can handle many data types, are resilient in the case of null values, are fast to train and can be interpreted at least globally.

\textbf{Deep Tabular Models}.
Due to the success of \acrshort{dl} in task involving texts, sound or images, many efforts are being made to find the best approach to apply these models to tabular data \cite{huang2020tabtransformer, gorishniy2021revisiting, somepalli2021saint, arik2021tabnet, joseph2022gate, kotelnikov2022tabddpm}. Most of these efforts belong to one of the 3 categories described below.

\textit{Modeling of multiplicative interactions between features}
Modeling explicitly the interaction between features of a tabular dataset \cite{he2017neural, guo2017deepfm, cheng2016wide, naumov2019deep, wang2021dcnv2} has been shown to have a significant impact on the performance of deep learning models in applications such as recommender systems and click-through-rate prediction. Nevertheless, recent comparisons \cite{borisov2022deep, gorishniy2021revisiting} show that these approaches produce worse outcomes than the rest of categories described below. 

\textit{Hybrid models}. 
Hybrid models transform the tabular data and combine deep neural networks with classical \acrshort{ml} approaches, frequently decision trees. Such hybrid models can be designed to be optimized in a fully-differentiable end-to-end or to benefit from non-differentiable approaches combined with deep neural networks. \acrshort{node} \cite{popov2019neural} is partially inspired by \acrshort{catboost} \cite{prokhorenkova2018catboost} and provides an example of fully differentiable model based on an ensemble of oblivious decision trees \cite{langley1994oblivious}. Entmax transformation and soft splits allow to obtain a fully differentiable end-to-end optimization. 
Other examples of fully-differentiable hybrid architecture are \cite{frosst2017distilling, luo2021sdtr, katzir2021netdnf}. On the other hand, DeepGBM model \cite{ke2019deepgbm} is an example of how to take advantage from the combination of non-differentiable approaches with deep neural networks. It combines deep neural network flexibility with gradient boosting decision tree preprocessing capabilities. TabNN \cite{ke2019tabnn} first distills the knowledge from gradient boosting decision trees to retrieve feature groups and then constructs a neural network based on feature combinations produced by clusterizing the results of the previous step.

\textit{Transformer-based models}. 
Many of \acrshort{dl} recent successes have been driven by the use of transformer-based methods \cite{devlin2019bert, radford2018improving, dosovitskiy2021an} inspiring the proposal of multiple approaches using deep attention mechanisms \cite{vaswani2017attention} for heterogeneous tabular data. 
The TabNet \cite{arik2021tabnet} design is inspired by decision trees: a set of subnetworks is processed in a hierarchical order and the results of all decision steps are aggregated in order to obtain the final prediction.
A feature transformer module chooses which features should be transferred to the next decision step and which should be employed to get the output at the present decision phase. 
TabTransformer \cite{huang2020tabtransformer} uses Transformer to improve the contextual embeddings of tabular features. First, each categorical variable goes through a specific embedding layer. A stack of Transformers is then used to enhance the categorical feature representation. The final contextual embedding is given by the concatenation of the so obtained categorical representation and the initial continuous features. 
In \acrshort{fttransformer} \cite{gorishniy2021revisiting}, columnar transformations (embeddings) are applied to both categorical and continuous features.  As in \acrshort{bert} \cite{devlin2019bert}, a \acrshort{cls} token is added to the set of columnar embeddings and then, a stack of transformer layers, are applied. The final \acrshort{cls} representation is employed as final contextual embedding, i.e. for predictions.   
\acrshort{saint} \cite{somepalli2021saint} combines the self-attention between features of the same tabular row with inter-sample attention over multiple-rows. When handling missing or noisy data, this mechanism allows the model to borrow the corresponding information from similar samples.  

As in \cite{huang2020tabtransformer, gorishniy2021revisiting, somepalli2021saint}, we investigate how contextual embedding affects the final model performance on supervised tasks. The main difference from the existing research is that in our approach, the contextual embedding is provided via \acrshort{gnn}s and, more specifically, by \acrshort{in}s.

\textbf{\acrlong{gnn} and \acrlong{in}}.
In case of neural networks such as Convolutional Neural Network or Transformer, the inputs must be structured data (grid and sequence, respectively). \acrshort{gnn} are a special subset of neural networks that can cope with less structured data, such as a graph. This means that the input can have arbitrary shapes and sizes, and can have complex topological relations. Permutation invariance is a crucial feature distinguishing \acrshort{gnn} from the rest of neural networks. The order of nodes in a graph has no relevance, this means, that the way in which we order the nodes in a graph does not impact the results produced by \acrshort{gnn}s. In a tabular dataset, the order of features (columns) does not have any meaning, so \acrshort{gnn} is a good candidate to model the interaction between them. 

The flow of a \acrshort{gnn} can be modeled using the Message-Passing scheme. a) For each pair of nodes $(u,v)$ in the graph, a message $M(u,v,e_{u,v})$ from $v$ to $u$ is created. Here $u$, $v$ are the embedding of nodes and $e_{u,v}$ is the (optional) embedding of edge. b) Each node aggregates the messages coming from all its neighbors. The aggregation must be permutation-invariant. c) The node is updated using its initial representation and the information obtained in point b. 

It is simple to find a map between the Message-Passing scheme and the contextual embedding of tabular features. a) Initial node representation is given by columnar feature embeddings. b) Message-passing through edges is the pairwise interaction between features. c) The neighbor aggregation represents the effect of the interaction of current feature with all its neighbors. d) The update step provides the contextual representation of each feature. 

In this paper, we investigate the benefits of using \acrshort{in}s for contextual embeddings of tabular data. They are a low-biased family of \acrshort{gnn} that have obtained enormous success when applied to simulation of complex physics or weather forecasting \cite{lam2022graphcast}.

The potential of \acrshort{gnn}s has attracted the community interest, and various attempts have been made to apply this type of solution to tabular data. To the best of our knowledge, past research has mostly focused on utilizing \acrshort{gnn} to learn relationships between samples in the same table or in distinct entities of a relational database. On the contrary, in our method we prioritize modeling feature relationships. The approaches are complimentary, and we leave it to future research to figure out how to integrate them. 

TabGNN \cite{guo2021tabgnn} focuses on modeling the relation between samples of the same table. Using a set of heuristics, a multiplex graph (i.e. a graph modeling different types of relations between nodes) is previously built from sample features. A specific \acrshort{gnn} obtains a customized sample representation for each edge type (i.e. for each type of node-to-node relation) and then an attention mechanism combines all contributions. This mechanism can be used in conjunction with other embedding strategies. In \cite{du2022learning}, to model the cross-sample and cross-column patterns a hypergraph is built from relevant data instance retrieval. Then a novel architecture of message-passing enhances the target data representation. Finally, in \cite{cvitkovic2020supervised, bai2021atjnet}, \acrshort{gnn}s are used to automatize and improve the features extraction in a relational database with a set of tables and foreign keys relationships. 

\section{\acrlong{ince}}
\label{sec:ince}
This section introduces the \acrshort{ince} model and describes its components in depth.

\textbf{Problem Definition}. We focus on supervised learning problems with tabular datasets $D = \left\lbrace x_i^{j_c}, x_i^{j_n}, y_i \right\rbrace_{i=1}^{N}$ where $x_i^{j_n}$ with $j_{n} \in [1, M_{num}]$ is the set of numerical features, $x_i^{j_c}$ with $j_{c} \in \left[1, M_{cat}\right]$ is the set of categorical features, $y_i$ is the label, $i \in \left[1, N\right]$ counts the dataset rows, $N$ is the total number of rows and $M = M_{num} + M_{cat}$ is total number of features.  

\textbf{Encoder-Decoder Perspective}.
\begin{figure}[ht]
	\centering
	\includegraphics[width=\linewidth]{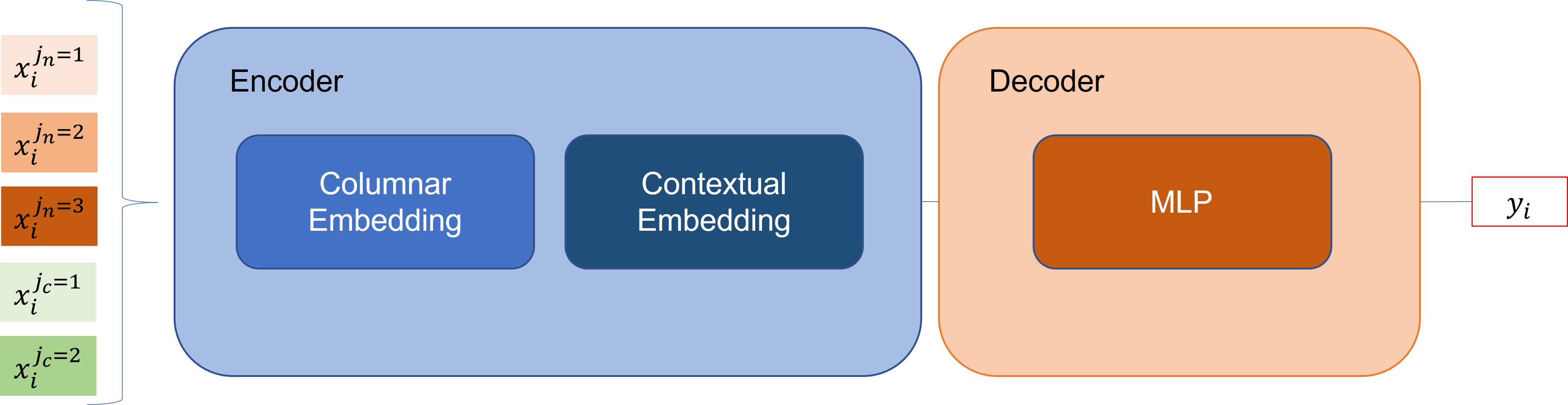}
	\caption{The encoder-decoder perspective \cite{hamilton2020graph}: an encoder model maps each tabular dataset feature into a latent vector, a decoder model uses the embeddings to solve the supervised learning task. In the encoding step, first a \textit{columnar} embedding individually projects any feature in a common latent space and then a \textit{contextual} embedding improves these representations taking into account the relationships among features. The decoder \acrshort{mlp} transforms the \textit{contextual} embedding output in the final model prediction.}
	\label{fig:encoder_decoder_perspective}
\end{figure}
As in \cite{hamilton2020graph}, we use the encoder-decoder perspective, Fig. \ref{fig:encoder_decoder_perspective}. First an encoder model maps each tabular dataset feature into a latent vector or embedding and then a decoder model takes the embeddings and uses them to solve the supervised learning task.

The encoder model is composed by two components: the \textit{columnar} and the \textit{contextual} embedding. The decoder model is given by a \acrshort{mlp} tuned to the learning task to solve.

\textbf{Encoder - Columnar Embedding}.
\begin{figure}[ht]
	\centering
	\includegraphics[width=\linewidth]{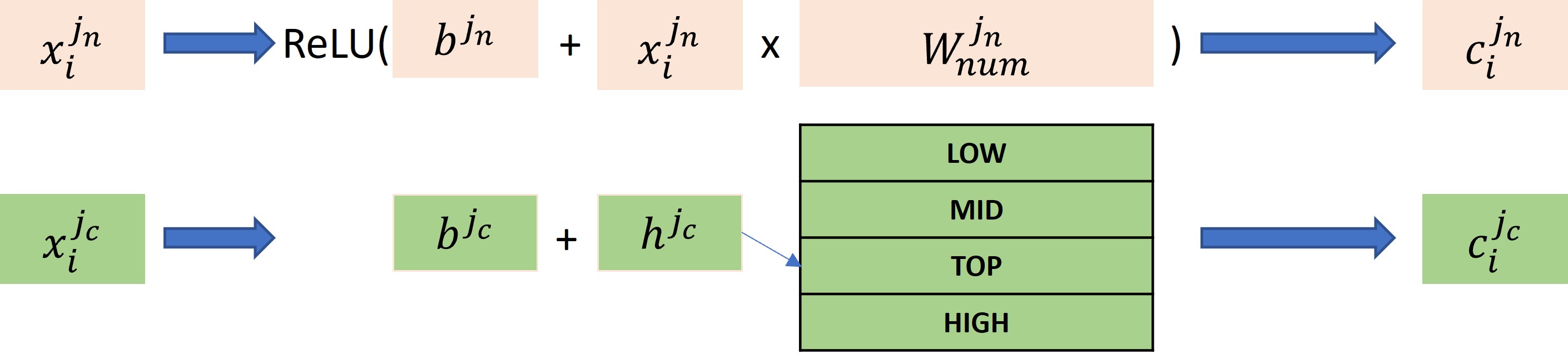}
	\caption{The \textit{columnar} embedding is responsible for projecting all the heterogeneous features in the tabular dataset in a common latent space. For each feature, a continuous or categorical transformation is defined. The \textit{columnar} embedding ignores any potential relationship or similarity between the tabular dataset features.}
	\label{fig:columnar_embedding}
\end{figure}
All of the original tabular heterogeneous features are projected in the same homogeneous and dense $d$-dimensional latent space by the \textit{columnar} embedding depicted in Fig. \ref{fig:columnar_embedding}. As in the \cite{gorishniy2021revisiting, somepalli2021saint}, the columnar embedding $c_i^{j_n}, c_i^{j_c} \in \mathbb{R}^d$ of continuous and categorical features $x_i^{j_n}, x_i^{j_c}$ are obtained as follows:
\begin{align}
c_{i}^{j_n} & = \text{ReLU}\left(b^{j_n} + x_{i}^{j_n} \cdot W^{j_n}_{num}\right)  &  W^{j_n}_{num} \in \mathbb{R}^d \\
c_{i}^{j_c} & = b^{j_c} + h_{j_c}^{T} W^{j_c}_{cat}  &  W^{j_c}_{cat} \in \mathbb{R}^{|j_c| \times d}
\end{align} 
where $\text{ReLU}$ is the non-linear activation function for the continuous embedding, $b^{j_n}$, $b^{j_c}$ are the feature bias, $W^{j_n}_{num} \in \mathbb{R}^d$ is a learnable vector, $W^{j_c}_{cat} \in \mathbb{R}^{|j_c| \times d}$ is a learnable lookup table and $|j_c|$ and $h_{j_c}^T$ are the size and the one-hot representation of the categorical feature $x_i^{j_c}$, respectively.   

\textbf{Encoder - Contextual Embedding}.
\begin{figure}[ht]
	\centering
	\includegraphics[width=\linewidth]{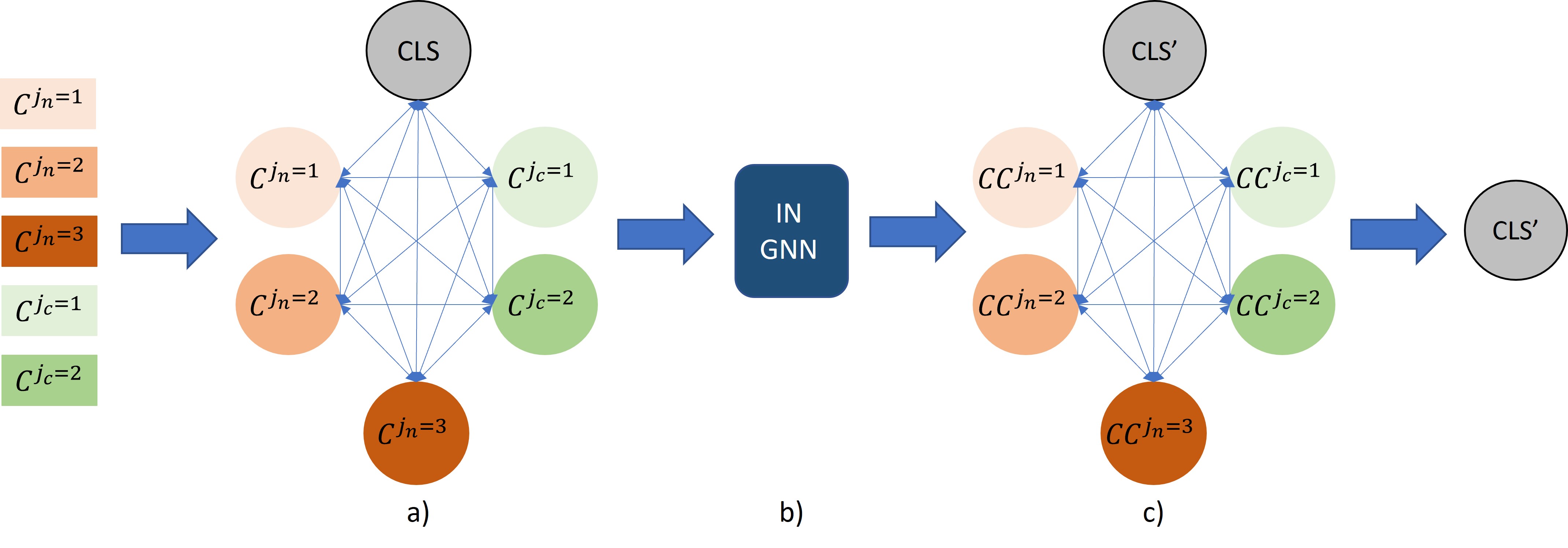}
	\caption{\textit{Contextual} embedding. (a) Homogeneous and fully-connected graph: it contains a node for each initial tabular features and a bidirectional-edge for each pair of nodes. The initial node representation is obtained by the \textit{columnar} embedding. A virtual \acrshort{cls} node is introduced to characterize the global graph state. (b) A stack of \acrshort{in} \cite{battaglia2016interaction} models node interactions to create a more accurate representation of nodes (i.e. tabular features). (c) The final representation of the \acrshort{cls} virtual node is used as \textit{contextual} embedding.}
	\label{fig:contextual_embedding}
\end{figure}
The \textit{columnar} embedding works feature by feature and has trouble identifying correlation or more general relationships between features in tabular datasets. To overcome this limitation, a \textit{contextual} embedding is introduced. In contrast to recent research \cite{huang2020tabtransformer, somepalli2021saint, gorishniy2021revisiting, arik2021tabnet} that use Transformers, we propose a \textit{contextual} embedding based on \acrshort{gnn} and, more specifically, \acrshort{in} \cite{battaglia2016interaction, battaglia2018relational, sanchezgonzalez2020learning}.

In this approach, the initial supervised learning task on tabular data is turned into a graph state estimation issue in which a categorical (classification task) or a continuous (regression task) graph state must be predicted. Taking into account the initial node representation (i.e. \textit{columnar} embedding) and graph edges, a stack of \acrshort{gnn} has to model the interactions among nodes in the latent space and learn a richer representation of the entire graph capable of improving state estimation. 

As shown in Fig. \ref{fig:contextual_embedding}, the first step consists of building a fully-connected graph. For each original tabular feature, a node is created $n_j \equiv x_j$ and for each pair of nodes $(n_{j_1}, n_{j_2})$, two directed and independent edges are defined: $e_{j_{1} j_{2}}: n_{j_1} \rightarrow n_{j_2}$ and $e_{j_{2} j_{1}}: n_{j_2} \rightarrow n_{j_1}$. The dense $d-$dimensional vector $c_j \in \mathbb{R}^d$ obtained from the \textit{columnar} embedding is used as initial node representation, giving rise to an homogeneous graph. No positional embedding is used to improve the node representation: the original tabular features are heterogeneous and each one is projected in the common latent space using a separate \textit{columnar} embedding. This is enough to distinguish the nodes among them without explicitly modeling their position in the graph\footnote{We have explicitly tested this hypothesis and the experiments confirm that the use of positional embedding does not improve the model performance.}. As in the \acrshort{bert} \cite{devlin2019bert}, a virtual \acrshort{cls}  node connected to each existing node is added to the graph. The $d-$dimensional initial representation of the \acrshort{cls} virtual node is a vector of learnable parameters. No features are initially considered for the edges $e_{ij}$. 

In the following step, a stack of \acrshort{in} is used to improve the representation of each node and edge in the graph. The final \acrshort{cls} vector embedding produced by the stack of \acrshort{in} is used as global representation of the graph, i.e. as a \textit{contextual} embedding of the tabular row\footnote{We have explicitly examined several approaches of pooling the node representation learned by \acrshort{gnn}. Our findings are consistent with the literature: the additional virtual \acrshort{cls} node method  outperforms all the other proposals.}.

\textbf{\acrlong{in}}.
\begin{figure}[ht]
	\centering		
	\includegraphics[width=\linewidth]{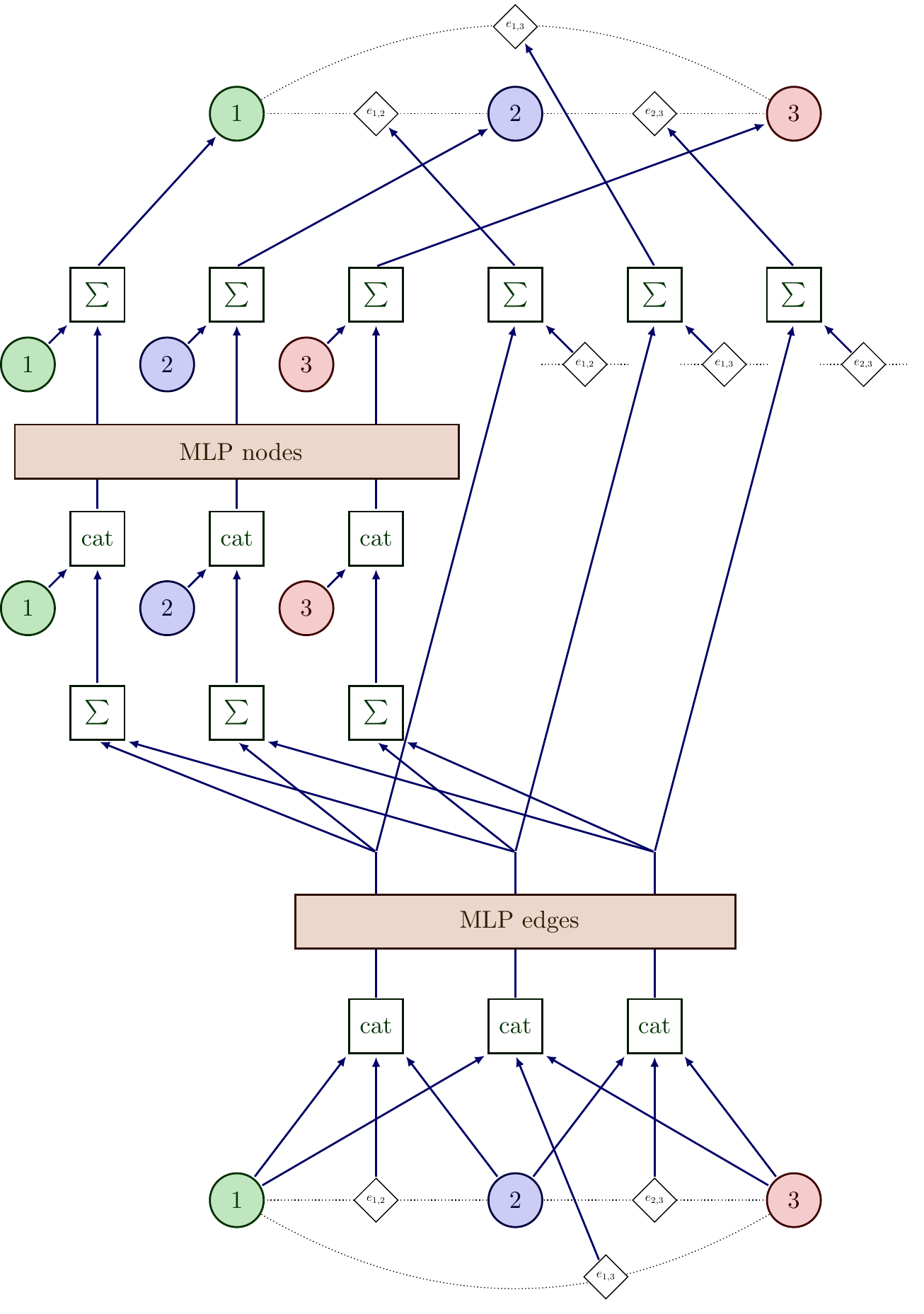}
	\caption{\acrlong{in} layer}
	\label{fig:ingnn}
\end{figure}
The workflow of a standard \acrshort{in} layer \cite{battaglia2016interaction, battaglia2018relational} is described in the Fig. \ref{fig:ingnn}. In the first step, the representation of each edge (i.e. interaction between each pair of tabular features) is updated using the information of the adjacent nodes (i.e. pair of tabular features):

\begin{equation}
e_{j_1 \rightarrow j_2}' = \text{\acrshort{mlp}}_{\text{E}} \left( \text{Concat} \left( n_{j_1}, n_{j_2}, e_{j_1 \rightarrow j_2}\right)\right) \,,
\label{eq:mlp_e}
\end{equation}
where $n_{j}, \, e_{j_1 j_2} \in \mathbb{R}^d$ are respectively node and edge representation, $\text{\acrshort{mlp}}_{\text{E}}$ is the shared neural network used to update all the graph edges. To simplify the notation we have suppressed the row index.

In the second step, all the messages coming from the incoming edges are aggregated and used to update the node representation:
\begin{equation}
n_{j}' = \text{\acrshort{mlp}}_{\text{N}} \left( \text{Concat} \left( n_{j},  \sum_{k \in \mathcal{N}} e_{k \rightarrow j}\right)\right) \,,
\label{eq:mlp_n}
\end{equation}
where $\mathcal{N}$ is the set of $n_{j}$ neighborhoods and $\text{\acrshort{mlp}}_{\text{N}}$ is the shared neural network used to update all the graph nodes. 

The residual connection between the initial and updated representations yields the final node and edge representations:  
\begin{align}
n_{j} & = n_{j}' + n_{j} \nonumber\\
e_{j_1 \rightarrow j_2} & = e_{j_1 \rightarrow j_2}' + e_{j_1 \rightarrow j_2} \,.
\label{eq:edge_update}
\end{align}

\textbf{Decoder}. The decoder $\text{\acrshort{mlp}}_{\text{DEC}}$ receives the contextual embedding computed by the encoder. It is a \acrshort{mlp} where the final output layer size and activation function are adapted to the supervised learning problem to solve - classification or regression.

\section{Experiments}
\label{sec:experiments}

\cite{borisov2022deep} provides a detailed review on the literature of \acrshort{dl} on tabular data together with an extensive empirical comparison of traditional \acrshort{ml} methods and \acrshort{dl} models on multiple real-world heterogeneous tabular datasets.

We consider the standard and deep models analyzed in \cite{borisov2022deep} as baseline and evaluate \acrshort{ince} using the tabular benchmark presented therein.  

\textbf{Data}.
The main properties of datasets are summarized in Table \ref{tab:datasets}.
\begin{table*}
	\caption{Tabular benchmark properties}
	\label{tab:datasets}
	\begin{center}
	\begin{tabular}{lcccc}
		\toprule
		\textbf{Dataset} & \textbf{Rows} & \textbf{Num. Feats} & \textbf{Cat. Feats} & \textbf{Task} \\ 
		\midrule
		\acrshort{heloc} & 9871 & 21 & 2 & Binary \\
		Cal. Hous. & 20640 & 8 & 0 & Regression \\
		Adult Inc.& 32561 & 6 & 8 & Binary \\
		Forest Cov. & 581 K & 10 & 2 (4 + 40) & Multi-Class (7) \\
		HIGGS & 11 M & 27 & 1 & Binary \\
		\bottomrule
	\end{tabular}
	\end{center}
\end{table*}

\textit{\acrshort{heloc}} \cite{heloc}: \acrfull{heloc} provided by FICO (a data analytics company), contains anonymized credit applications of \acrshort{heloc} credit lines.  The dataset contains 21 numerical and two categorical features characterizing the applicant to the \acrshort{heloc} credit line. The task is a binary classification and the goal is to predict whether the applicant will make timely payments over a two-year period. 

\textit{California Housing} \cite{californiahousing}: The information refers to the houses located in a certain California district, as well as some basic statistics about them based on 1990 census data. This is a regression task, which requires to forecast the price of a property. 

\textit{Adult Incoming} \cite{adult}: Personal details such as age, gender or education level, are used to predict whether an individual would earn more or less than $50K\$$ per year.

\textit{Forest Cover Type} \cite{covertype}: Cartographic variables are used to predict the forest cover type: it is a multi-class (seven) classification task. The first eight features are continuous whereas the last two are categorical, with four and 40 levels respectively.

\textit{HIGGS} \cite{higgs}: The dataset contains 11M of rows and 28 features where the first 21 are kinematic properties measured by the particle detectors, and the last seven are processed features built by physicists. The data has been produced using Monte Carlo simulations and the binary classification task is to distinguish between signals with Higgs bosons and a background process.

\textbf{Data Preprocessing}. In order to compare \acrshort{ince} with the results of \cite{borisov2022deep}, we reproduce the same data preprocessing. Zero-mean and unit-variance normalization is applied to the numerical features whereas an ordinal encoding is used for the categorical ones. The missing values were imputed with zeros.

\textbf{Baselines}. \acrshort{ince} is compared to the following models. \textit{Standard methods}: Linear Model, KNN, Decision Tree, Random Forest \cite{breiman2001randomforests}, \acrshort{xgboost} \cite{chen2016xgboost}, \acrshort{lightgbm} \cite{ke2017lightgbm}, \acrshort{catboost} \cite{prokhorenkova2018catboost}. \textit{Deep learning models}: \acrshort{mlp} \cite{mcculloch1943logical} , DeepFM \cite{guo2017deepfm}, DeepGBM \cite{ke2019deepgbm}, RLN \cite{shavitt2018regularization}, TabNet \cite{arik2021tabnet}, VIME \cite{yoon2020vime}, TabTrasformer \cite{huang2020tabtransformer}, \acrshort{node} \cite{popov2019neural}, \acrshort{netdnf} \cite{katzir2021netdnf}, \acrshort{saint} \cite{somepalli2021saint}, \acrshort{fttransformer} \cite{gorishniy2021revisiting}.

\textbf{Setup}. For each tabular dataset, we use the Optuna library \cite{akiba2019optuna} with 50 iterations to tune \acrshort{ince} hyperparameters. Each hyperparameter configuration is cross-validated with five folds. The search space is the following: latent space size $\in \left\lbrace 32, 64, 128 \right\rbrace $, number of stacked \acrshort{in} $ \in \left\lbrace 1, 2, 3, 4\right\rbrace $ and depth of $\text{\acrshort{mlp}}_{\text{E}}, \, \text{\acrshort{mlp}}_{\text{N}} \in \left\lbrace 1, 2, 3, 4\right\rbrace $.

In all the experiments, we consider a decoder $\text{\acrshort{mlp}}_{\text{DEC}}$ with two hidden layers and $\text{ReLU}$ is the non-linear activation function used for  $\text{\acrshort{mlp}}_{\text{E}}$, $\text{\acrshort{mlp}}_{\text{N}}$ and $\text{\acrshort{mlp}}_{\text{DEC}}$. Cross-Entropy and \acrfull{mse} are the loss functions used in classification and regression tasks, respectively. We train all the models $200$ epochs using Adam optimizer with a learning rate of $0.001$ and with batches of size $256$. All the \acrshort{dl} code is implemented using PyTorch \cite{paszke2019pytorch} and PyTorch-Geometric \cite{fey2019fast} and parallelized with Ray \cite{moritz2018ray}. 

\subsection{Results}
\label{subsec:results}

\begin{figure}[ht]
	\centering
	\includegraphics[width=\linewidth]{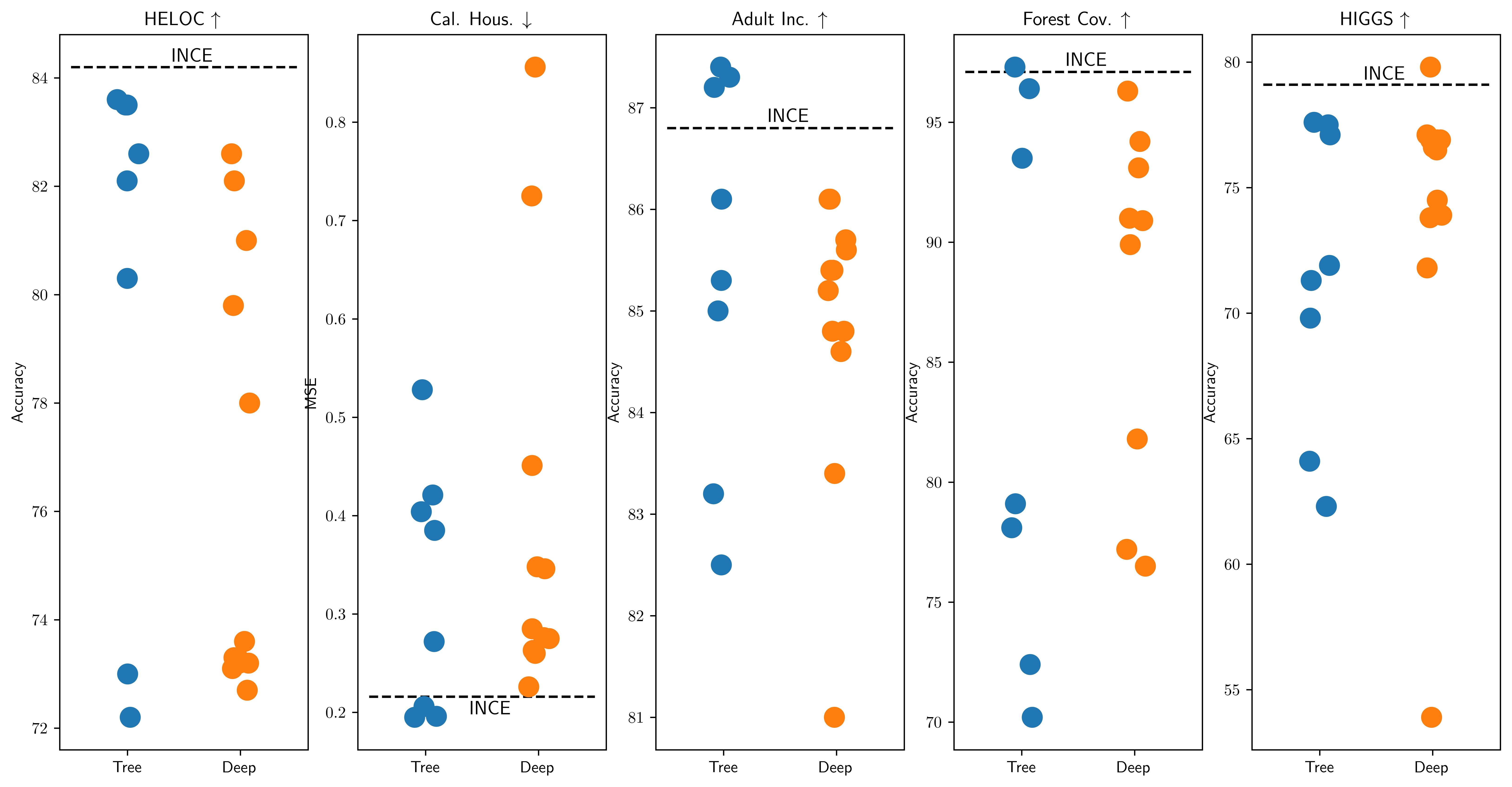}
	\caption{The stripplot in blue and orange illustrate the distribution of tree-based and \acrshort{dl} baseline, respectively. The horizontal dotted line represents the \acrshort{ince} performance. Accuracy and \acrshort{mse} are the metrics used for classification and regression tasks. The presence of an up/down arrow near the dataset name indicates whether the metric must be maximized o minimized.}
	\label{fig:boxplot_results}
\end{figure}

In Table \ref{tab:results} and Fig. \ref{fig:boxplot_results}, we report the results on the tabular benchmark described above. In four of five datasets, \acrshort{ince} outperforms all the \acrshort{dl} baselines. In the fifth, HIGGS case, \acrshort{ince} obtains the second best performance behind \acrshort{saint} model \cite{somepalli2021saint}, but largely above the rest of \acrshort{dl} models. In two of the five datasets, \acrshort{ince} outperforms tree-based models, while in the other three it achieves results that are competitive with them.

In terms of baseline performance, we carefully reproduced the findings for \acrshort{xgboost}, \acrshort{mlp}, TabTransformer, and \acrshort{saint} to ensure that our preprocessing and optimization approach was equivalent to \cite{borisov2022deep} for all datasets in the benchmark. After demonstrating the comparability of \cite{borisov2022deep} and our flows, the other baseline results are quoted from this paper. It should be noted that we include in our study the \acrshort{fttransformer} that is subsequent to \cite{borisov2022deep}.

\begin{table*}
	\caption{\acrshort{ince} model vs. Best Tree-based model vs. Best Deep model }
	\label{tab:results}
	\begin{center}
		\begin{tabular}{cccccccccc}
			\toprule			
			\multirow{2}{*}{\textbf{Dataset}} & \multirow{2}{*}{\textbf{Metrics}} & \multicolumn{2}{c}{\textbf{Best Tree}} & & \multicolumn{2}{c}{\textbf{Best \acrshort{dl}}} & & \multicolumn{2}{c}{\textbf{\acrshort{ince}}} \\
			\cmidrule{3-4}\cmidrule{6-7}\cmidrule{9-10}
			& & \textit{Result} & \textit{Model} && \textit{Result} & \textit{Model} && \textit{Result} & \textit{Rank}\\
			\midrule
			\acrshort{heloc} & Accuracy $\uparrow$ & $83.6\%$ & \acrshort{catboost} && $82.6\%$ & \acrshort{netdnf} && $84.2 \pm 0.5\%$ & 1st Abs. \\
			Cal. Hous. & \acrshort{mse} $\downarrow$ & $0.195$ & \acrshort{lightgbm} && $0.226$ &  \acrshort{saint} && $0.216 \pm 0.007$ & 1st \acrshort{dl} \\
			\multirow{2}{*}{Adult Inc.} & \multirow{2}{*}{Accuracy $\uparrow$} &  \multirow{2}{*}{$87.4\%$} & \multirow{2}{*}{\acrshort{lightgbm}} && \multirow{2}{*}{$86.1\%$} 
			& DeepFM && \multirow{2}{*}{$86.8 \pm 0.3\%$} & \multirow{2}{*}{1st \acrshort{dl}} \\
			&&&&&& \acrshort{saint} &&& \\
			\multirow{2}{*}{Forest Cov.} & \multirow{2}{*}{Accuracy $\uparrow$} & \multirow{2}{*}{$97.3\%$} & \multirow{2}{*}{\acrshort{xgboost}} && \multirow{2}{*}{$96.3\%$} & \multirow{2}{*}{\acrshort{saint}} && \multirow{2}{*}{$97.1 \pm 0.1\%$} & 1st \acrshort{dl} \\
			& & & & & & &&& 2nd Abs. \\
			\multirow{2}{*}{HIGGS} & \multirow{2}{*}{Accuracy $\uparrow$} & \multirow{2}{*}{$77.6\%$} & \multirow{2}{*}{\acrshort{xgboost}} && \multirow{2}{*}{$79.8\%$} & \multirow{2}{*}{\acrshort{saint}} && \multirow{2}{*}{$79.1 \pm 0.0\%$} & 2nd \acrshort{dl} \\
			& & & & & & & && 2nd Abs. \\
			\bottomrule
		\end{tabular}
	\end{center}
	\begin{tablenotes}
		\RaggedRight
		\item The Accuracy and \acrshort{mse} are the metrics used for classification and regression tasks, respectively. The presence of an up/down arrow near the name indicates whether the metric must be maximized o minimized. For \acrshort{ince}, the mean and standard deviation are reported together with its ranking.
	\end{tablenotes}
\end{table*}

\section{Deep Dive in \acrlong{in}}

For each tabular dataset, we have studied how the choice of latent space size $l$, $\text{\acrshort{mlp}}_{\text{N, E}}$ depth $d$ and number $n$ of stacked \acrshort{in} influences the model behavior: number of trainable parameters, performances and computational time. The findings from the various datasets reveal similar patterns, leading to consistent conclusions.

\textbf{Trainable parameters}. The number of trainable parameters $\mathcal{TP} \left( \text{\acrshort{in}}\right)$ of a stack of $n$ \acrshort{in} is given by: 
\begin{equation}
\label{eq:in_learnable_params}
\begin{split}
\mathcal{TP} \left( \text{\acrshort{in}}\right)& = \sum_{i=1}^{n} \mathcal{TP} \left( \text{\acrshort{in}}^{i}\right)  \\
& = \sum_{i=1}^{n} \left[\mathcal{TP}\left(\text{\acrshort{mlp}}_{\text{E}}^{\text{i}} \right) + \mathcal{TP}\left( \text{\acrshort{mlp}}_{\text{N}}^{\text{i}} \right)\right] \\
\mathcal{TP}\left(\text{\acrshort{mlp}}_{\text{N}}^{\text{i}} \right) & = \left( 2 \cdot l^2 + l \right) + \left( d - 1 \right) \cdot \left( l^2 + l \right)  \\
\mathcal{TP}\left( \text{\acrshort{mlp}}_{\text{E}}^{\text{i}} \right) & = \left(K_i \cdot l^2 + l \right) + \left(d - 1 \right) \cdot \left(l^2 + l \right) \,,
\end{split}
\end{equation}
where $K_i = 2$ if $i=1$ and $K_i = 3$ otherwise. We consider all the hidden layers of $\text{\acrshort{mlp}}_{\text{E, N}}$ of the same size. The difference in the number of parameters between $\text{\acrshort{mlp}}_{\text{E}}^{i=1}$ and $\text{\acrshort{mlp}}_{\text{E}}^{i>1}$ is due to the fact that all \acrshort{in} with $i > 1$ receive the edge features computed by preceding layers, whilst the first \acrshort{in} does not use any initial edge features.  

\begin{figure}[ht]
	\centering
	\includegraphics[width=\linewidth]{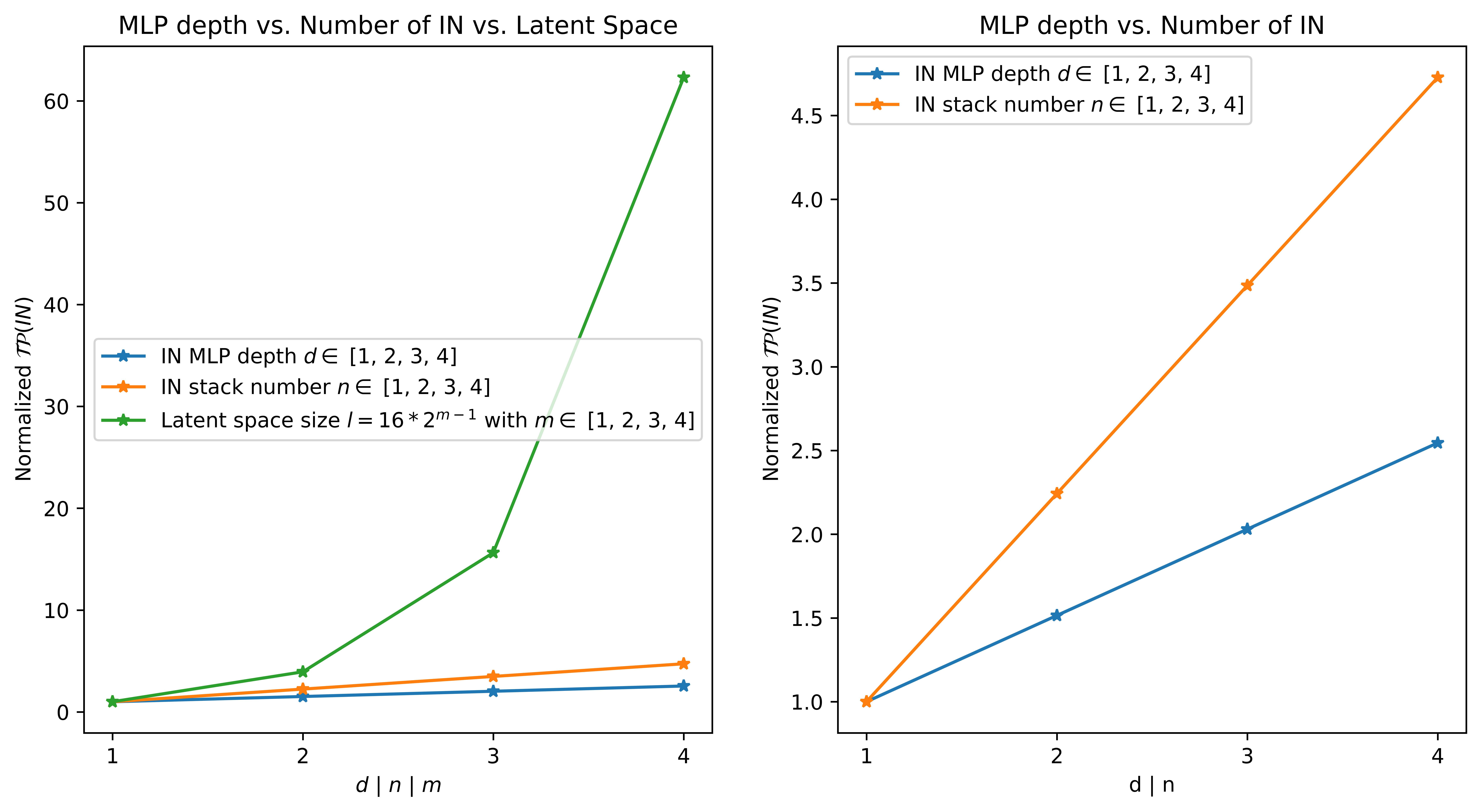}
	\caption{Growth of the normalized $\mathcal{TP} \left( \text{\acrshort{in}}\right)$ as a function of $\text{\acrshort{mlp}}_{\text{E, N}}$ depth, number of stacked $\text{\acrshort{in}}$ and latent space size. The plot on the left compares the evolution of $\mathcal{TP} \left( \text{\acrshort{in}}\right)$ when two hyperparameters are fixed and the third is increased. The plot on right is a zoom on the contribution of $\text{\acrshort{mlp}}_{\text{E, N}}$ depth and number of stacked $\text{\acrshort{in}}$. The baseline used to normalize $\mathcal{TP} \left( \text{\acrshort{in}}\right)$ is given by the number of trainable parameters of the simplest case: $l=16$, $d=1$, $n=1$. It is trivial to show using Eq. \ref{eq:in_learnable_params} that the behavior of normalized $\mathcal{TP} \left( \text{\acrshort{in}}\right)$ curve does not depend on the particular choice of the baseline latent space size $l$.}
	\label{fig:trainable_params}
\end{figure}

The quantity of trainable parameters increases quadratically with the size of the latent space and linearly with the number of stacked $\text{\acrshort{in}}$ or the $\text{\acrshort{mlp}}_{\text{E, N}}$ depth, Fig. \ref{fig:trainable_params}. The slope of the straight line corresponding to the number of stacked $\text{\acrshort{in}}$ is steeper than the one relative to the $\text{\acrshort{mlp}}_{\text{E, N}}$ depth.

\textbf{Performances}. Our experiments suggest that whereas the latent space size needs to be fine-tuned for each dataset, the impact of $\text{\acrshort{mlp}}_{\text{E, N}}$ depth $d$ and number $n$ of stacked $\text{\acrshort{in}}$ does not depend on the supervised learning problem to solve. The configuration with $d=3$ and $n=2$ is a solid baseline regardless of the underlying task.

\begin{figure}
	\centering
	\includegraphics[width=\linewidth]{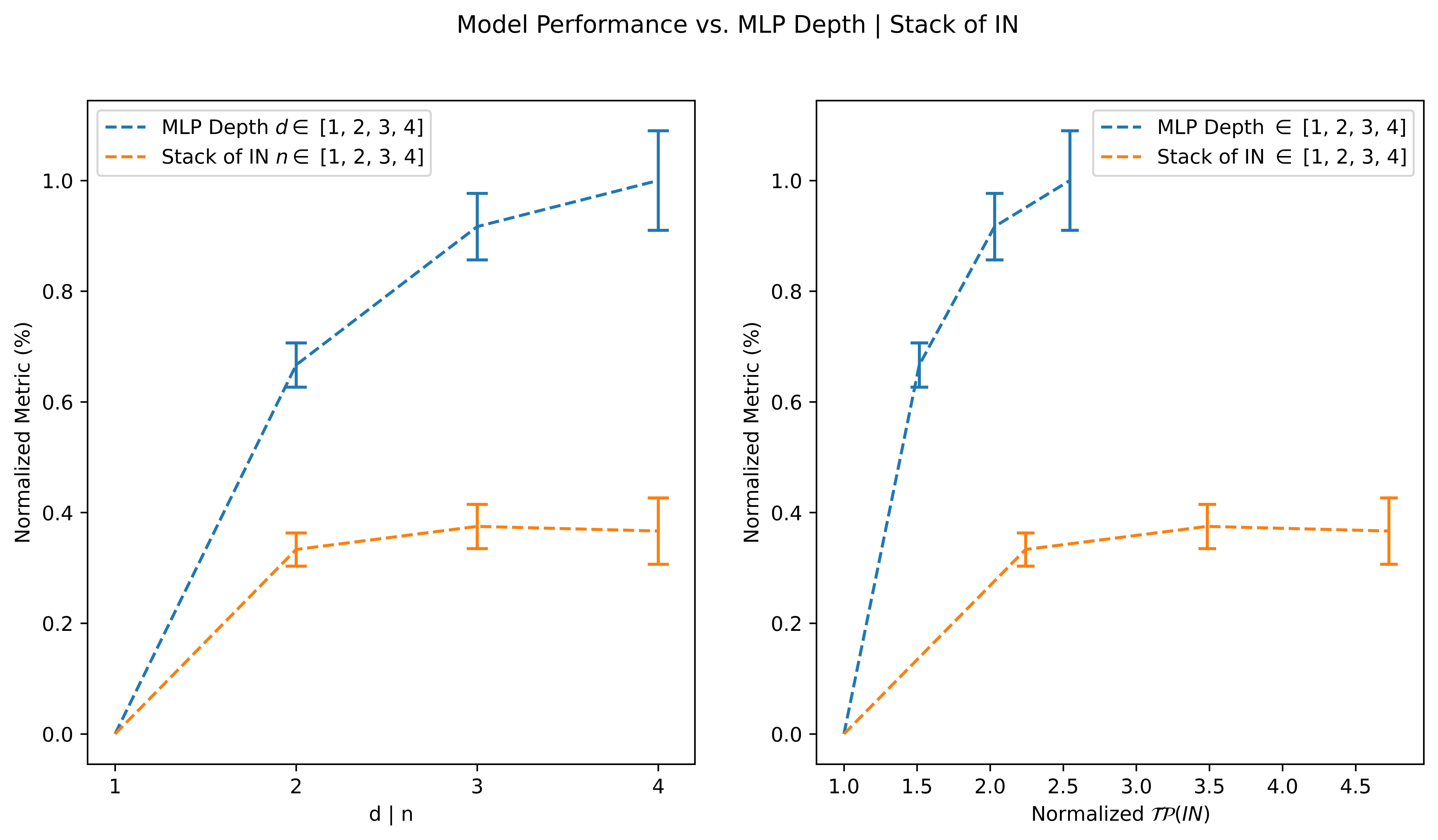}
	\caption{Average Normalized Metric. Left side plot depicts how the normalized metric changes when the $\text{\acrshort{mlp}}_{\text{E, N}}$ depth or the number of stacked $\text{\acrshort{in}}$ is increased and the other is kept constant. The right side plot shows the same information but referenced to the normalized number of trainable parameters.}
	\label{fig:performance_analysis}
\end{figure}

To clarify this point, in Fig. \ref{fig:performance_analysis} we show how the normalized metric changes as a function of the $\text{\acrshort{mlp}}_{\text{E, N}}$ depth and the number of stacked  $\text{\acrshort{in}}$. The normalized metric is a global performance measure (higher is better) generated using the findings from all of the datasets as described in Appendix \ref{sec:normalization_metric}.

The left side plot in Fig. \ref{fig:performance_analysis} depicts the normalized metric curves $\mathcal{C}_{d}$ (blue line) and $\mathcal{C}_{n}$ (orange line) obtained modifying $d$ and $n$ respectively while the other parameters are kept constant. The information on the right side plot is the same as on the left, but it is compared to the normalized number of trainable parameters. 

The depth $d$ of the shared neural networks $\text{\acrshort{mlp}}_{\text{N, E}}$ has the most impact on the model performances and, at the same time, it has reduced effect on the number of learnable parameters. These results are coherent with the observed behavior of the Optuna \cite{akiba2019optuna} bayesian optimizer. Regardless of the supervised learning problem, after few attempts, it quickly reduces search space for $d$ to $[3,4]$ and then it fine-tunes the number of stacked $\text{\acrshort{in}}$ in the range $[2, 3]$. The configuration with $d=3$ and $n=2$ is always a solid candidate regardless of the tabular dataset.

Why adding more than two layers does not improve the \textit{contextual} encoder capability? We interpret this as follows. a) The number of nodes in the graph is small. In our formulation there is a node for each tabular feature and the number of them goes from eight (California Housing) to 28 (HIGGS). After two \acrshort{in} layers, the information of a node has been transmitted to every other node in the graph. b) We are working with a fully connected graph, i.e. a trivial topology. The \acrshort{in} has to model the strength of each edge but the initial topological information seems to be poor. c) The size of datasets is limited (excluding HIGGS).

\textbf{Computational time}. 
\begin{figure}[ht]
	\centering
	\includegraphics[width=\linewidth]{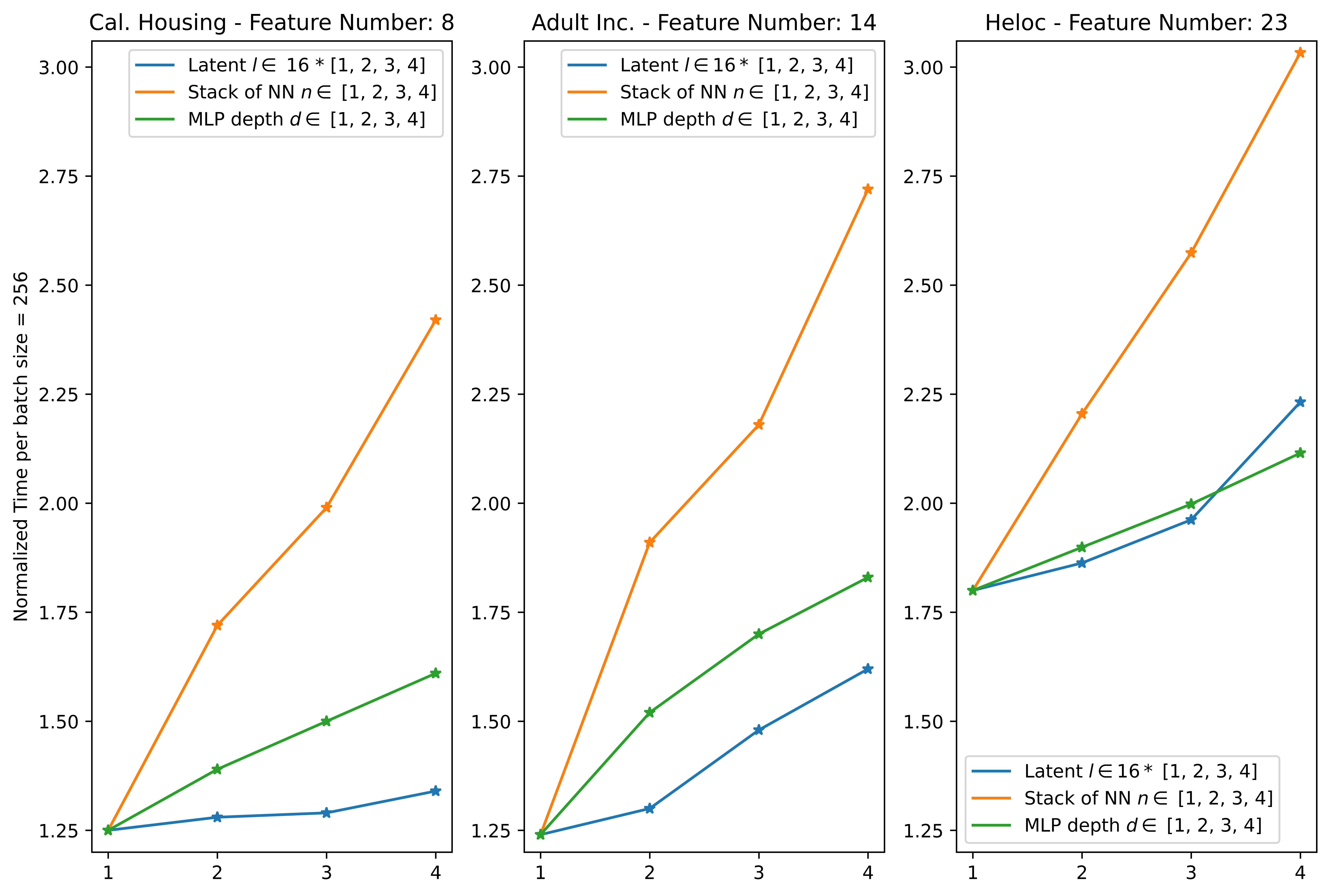}
	\caption{Average normalized training time. For each dataset the \acrshort{ince} training time is normalized using the time of the corresponding \acrshort{mlp} with the same columnar embedding and decoder but without contextual embeddings. All the results are relative to a batch size of $256$. Starting from the configuration base $l=16$, $n=1$ and $d=1$, the different curves are computed modifying one parameter while the others are kept constant.}
	\label{fig:time_ingnn}
\end{figure}
Fig. \ref{fig:time_ingnn} shows how the number of features in the tabular dataset as well as the \acrshort{ince} configuration (latent space size, number of stacked \acrshort{in} and $\text{\acrshort{mlp}}_{\text{N, E}}$ depth) impact on the training time.  In particular, Fig. \ref{fig:time_ingnn} presents the average traininig time for a batch size of $256$. All the \acrshort{ince} training times are normalized by using the corresponding train time of a \acrshort{mlp} with the same columnar embedding and the same decoder but without contextual embeddings. For each dataset, the three curves are obtained modifying one parameter (for example $n \in \left\{1, 2, 3, 4\right\}$ for the orange line) while holding the other two constant ($l=16$ and $d=1$).
\begin{itemize}
	\item As expected, the number of features in the tabular dataset has an effect on the computational time: it grows from California Housing (eight features) to Heloc ($23$ features) for a fixed \acrshort{ince} configuration. In our proposal, we are working with a fully-connected graph and the volume of operations increases quadratically regarding to the number of nodes (features). 
	\item For a fixed dataset, the number $n$ of stacked \acrshort{in} has the greatest impact on the amount of operations and, hence, on computational time.
	\item When the number of features is around $20$, the impact of latent space size is comparable or even greater than the impact of $\text{\acrshort{mlp}}_{\text{N, E}}$ depth.
\end{itemize}

\subsection{\acrlong{in} vs. Transformer}

Recent works \cite{huang2020tabtransformer, somepalli2021saint, gorishniy2021revisiting} propose the Transformers encoder \cite{vaswani2017attention} as \textit{contextual} embedding. Here, we analyze similarities and differences between the two approaches.

\textbf{Approach}. In this work, we concentrate on the use case where Transformer encoders or \acrshort{gnn}s are employed to learn the interaction between features improving the contextual embedding. For this particular use case, the following features are shared by both approaches: 
\begin{itemize}
	\item The columnar embeddings of each individual feature are organized in a fully-connected graph with an additional extra virtual node (\acrshort{cls}).
	\item A mechanism (the attention mechanism in the Trasformer case and the \acrshort{in} convolution of Eqs. \ref{eq:mlp_e}, \ref{eq:mlp_n} in our proposal) models the interaction between nodes/features. The strength of the interaction between nodes acts as a \textit{soft prune mechanism}: the stronger the interaction between a neighborhood with the current node, the larger its contribution to the current node contextual embedding.
\end{itemize}   
The main difference is in how the interaction is modeled. 
In the original Attention mechanism \cite{vaswani2017attention}, the contextual node embedding (for head=1) is given by (neglecting for sake of simplicity the skip connection in both cases, Transformer encoders and \acrshort{gnn}s):
\begin{equation}
n^{\prime\alpha}_{i} = \sum_{j=1}^{M} \sum_{\beta=1}^{l} \omega_{i,j} V^{\alpha, \beta} n^{\beta}_{j} \,,
\label{eq:attn1}
\end{equation}
where latin indexes $i,j=1,\dots,M$ are indexes in the topological space (that is over the graph nodes), Greek indexes $\alpha, \beta = 1, 2,\dots, l$ are indexes in the latent space and $\omega_{i,j}$ is the attention mechanism:
\begin{equation}
\omega_{i,j} = \text{softmax}_{j} \left( \dfrac{ n^{i} Q K^T n^{j}}{\sqrt{l}} \right) \,.
\label{eq:att2}
\end{equation} 
The Eq. \ref{eq:attn1} shows that the interaction between nodes $n_i$ and $n_j$ is written as the product of two operators $\omega_{i,j}$ and  $V^{\alpha, \beta}$. The attention mechanism $\omega_{i,j}$ is an operator with no trivial structure in the topological space (it depends on the nodes indexes $i$ and $j$) but diagonal in the latent space (it does not depend on the indexes in latent space). $V^{\alpha, \beta}$, on the contrary, is diagonal in the topological space (it does not depend on the node indexes) but with a non-trivial structure in the latent space (it depends on the latent space indexes $\alpha$, $\beta$). 

Using Eqs. \ref{eq:mlp_e}, \ref{eq:mlp_n}, it is possible to prove that, for the case of \acrshort{in}, the node contextual embedding is given by:
\begin{equation}
n^{\prime\alpha}_{i} = \text{\acrshort{mlp}}_{N} \left( \text{Concat} \left(n_i, \sum_j \text{\acrshort{mlp}}_E \left( n_i, n_j, e_{n_j \rightarrow n_i} \right) \right) \right) \,.
\end{equation} 
This is a more general formulation than the case of the attention mechanism. $\text{\acrshort{mlp}}_{N}$ plays a similar role to the $V^{\alpha, \beta}$ operator of the transformer encoder and does not depend on the node indexes. The $\text{\acrshort{mlp}}_E$ plays a similar role to $\omega_{i,j}$. The main difference is that the $\text{\acrshort{mlp}}_E$ is a non-trivial operator in both topological and latent space and therefore, given a pair of nodes $n_i$ and  $n_j$, it may learn different strengths for different latent space indexes $\alpha$, $\beta$.

\textbf{Performances}. For comparison purposes, we replace in our flow the \acrshort{in} with a Transformer encoder while keeping intact the rest of the model components - i.e. same \textit{columnar} embedding and same decoder. See \cite{vaswani2017attention} for the details about Transformer models and its components: Multi Head Self-Attention and FeedForward block. Using Optuna \cite{akiba2019optuna}, we look for the best set of Transformer encoder hyperparameters in the following search space: number of attention heads $h \in \left\lbrace 1, 2, 4, 8\right\rbrace $, FeedForward layer space size $f \in \left\lbrace 512, 1024, 2048\right\rbrace $, number of stacked Transformer encoders $n \in \left\lbrace 1, 2, 3, 4\right\rbrace $ and latent space size $l \in \left\lbrace 16, 32, 64, 128\right\rbrace $. As in the \acrshort{in} case, each hyperparameter configuration is cross-validated five folds and all the models are trained $200$ epochs using Adam optimizer with a learning rate of $0.001$ and  batches of size $256$.

Table \ref{tab:trnasf_in} shows how the two approaches provide comparable results even though, at least on the selected benchmark, the \acrshort{in} encoder performs slightly better.

\begin{table}[H]
	\caption{\acrshort{ince} vs. Transformer contextual embedding}
	\label{tab:trnasf_in}
	\begin{center}
		\begin{tabular}{cccc}
			\toprule			
			\textbf{Dataset} & \textbf{Metrics} & \textbf{Transformer} & \textbf{\acrshort{ince}} \\
			\midrule
			\acrshort{heloc} & Acc. $\uparrow$ & $83.8 \pm 0.6 \%$ & $84.2 \pm 0.5\%$\\
			Cal. Hous. & \acrshort{mse} $\downarrow$ & $0.228 \pm 0.006$ & $0.216 \pm 0.007$ \\
			Adult Inc. & Acc. $\uparrow$ & $86.5 \pm 0.3\%$ & $86.8 \pm 0.3\%$ \\
			Forest Cov. & Acc. $\uparrow$ & $95.8 \pm 0.1\%$ & $97.1 \pm 0.1\%$ \\
			HIGGS & Acc. $\uparrow$ & $78.5 \pm 0.0 \%$ & $79.1  \pm 0.0\%$ \\
			\bottomrule
		\end{tabular}
	\end{center}
\end{table}

\textbf{Trainable parameters}. The size $f$ of the latent space used by the Transformer FeedForward block has a significant impact on the number of trainable parameters in a Transformer encoder. In the comparison\footnote{For the purpose of simplicity, we exclude the Normalization Layers parameters from our study in both cases, Transformer and \acrshort{in}.} that follows, we take into account the setup where $f=512$ since it achieves the best average results in the Optuna optimization.

The number of trainable parameters of a Transformer encoder is given by: 
\begin{equation}
\label{eq:transformer_params}
\begin{split}
\mathcal{TP} \left(\text{Transformer}\right) & = n \cdot  \left[\right.
\mathcal{TP} \left(Q, K, V \right) + \\
&\text{\color{white} = } \; \; \; \; \; \; \; \; \mathcal{TP} \left(\text{MultiAttention} \right) + \\
&\text{\color{white} = }  \; \; \; \; \; \; \; \; \mathcal{TP} \left(\text{FeedForward} \right) \left.\right] \\ 
\mathcal{TP} \left(Q, K, V \right) & = 3 \cdot h \cdot l \cdot (l + 1)  \\
\mathcal{TP} \left(\text{MultiAttention} \right) &  = l \cdot ( h \cdot l + 1)  \\
\mathcal{TP} \left(\text{FeedForward} \right) & = 2 \cdot f \cdot l + f + l \,,
\end{split}
\end{equation}
where $l$, $h$, $f$ and $n$ are respectively the latent space size, the number of attention heads, the FeedForward latent space size and the number of stacked Transformer encoders. 

\begin{figure}[ht]
	\centering
	\includegraphics[width=0.75\linewidth]{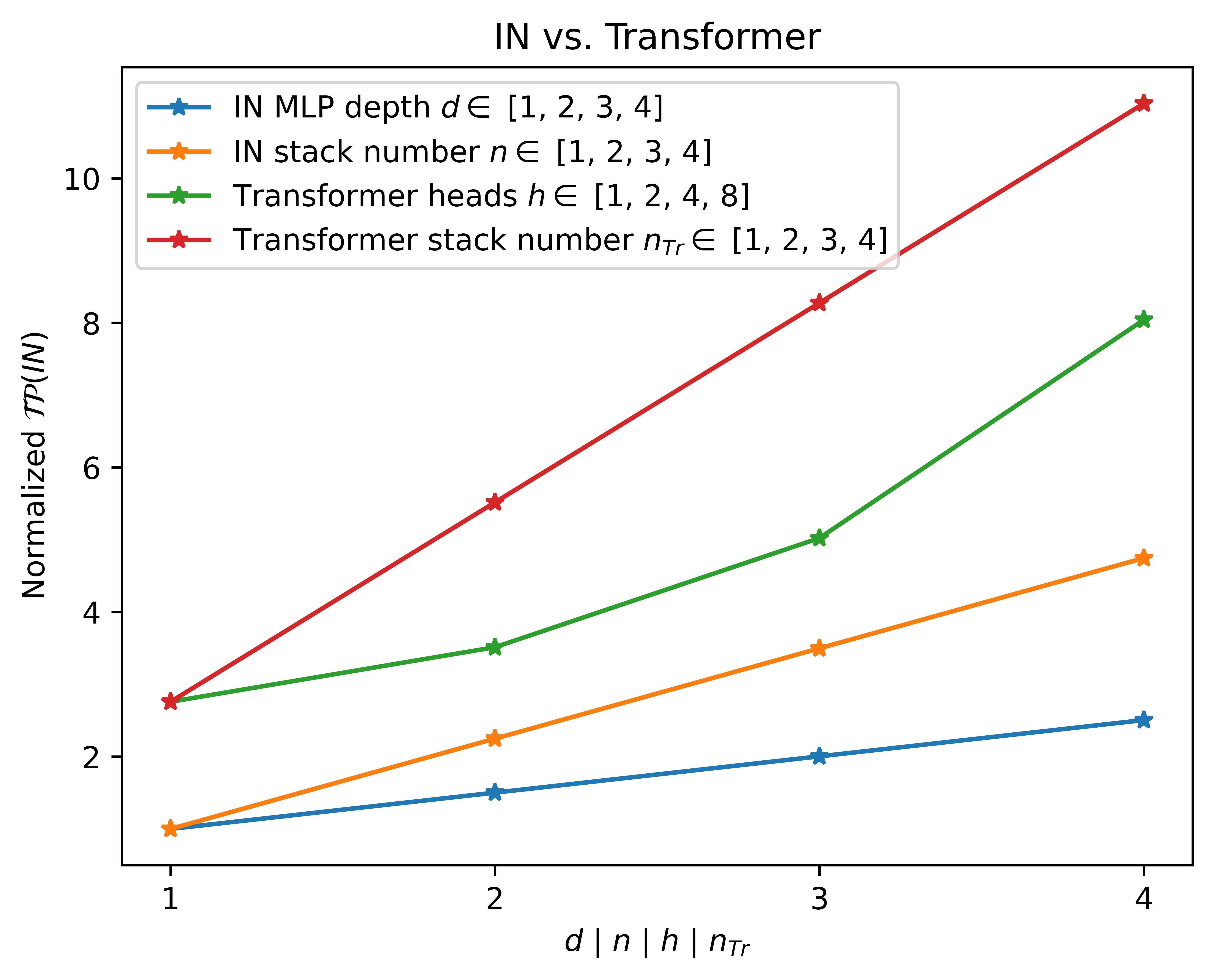}
	\caption{Comparison of \acrshort{in} and Transformer trainable parameters. The normalized $\mathcal{TP}$ is obtained using Eq. \ref{eq:in_learnable_params} and Eq. \ref{eq:transformer_params} and then normalizing with regarding to $\mathcal{TP} \left(\acrshort{in}_{l, d=1, n=1}\right)$. The plot shows the results for $l=128$. Transformers have more trainable parameters than \acrshort{in}, and the relative difference grows when $l$ decreases.}
	\label{fig:transformer_vs_in_weights}
\end{figure}

Fig. \ref{fig:transformer_vs_in_weights} compares the behavior of $\mathcal{TP} \left(\text{Transformer}\right)$ and  $\mathcal{TP} \left(\text{\acrshort{in}}\right)$. As in Fig. \ref{fig:trainable_params}, the normalized number of trainable parameters $\mathcal{TP}$ is obtained dividing by $\mathcal{TP} \left(\text{\acrshort{in}}_{l, d=1, n=1}\right)$. The Fig. \ref{fig:transformer_vs_in_weights} presents the results for $l=128$. \acrshort{in} has less trainable parameters than Transformers and the relative difference is even bigger when $l$ decreases. When the number of attention heads is $h \le 2$, the difference is due to the FeedForward block parameters. For $h > 2$, Transformer has more parameters included, without taking into account the FeedForward block.

\textbf{Limitations}. When the number of tabular features increases, both \acrshort{in} and Transformers use greater resources. The vanilla Multi Head Self-Attention and \acrshort{in} on fully-connected graph share quadratic complexity regarding the number of features. This issue can be mitigated by using efficient approximations of Multi Head Self-Attention \cite{tay2022efficient} or a more complex graph topology with less edges in the \acrlong{in} case. Additionally, it is still possible to distill the final model into simpler architectures for better inference performance.

\section{Interpretability of contextual embedding}

\subsection{Columnar vs. Contextual embedding}

In subsection \ref{subsec:results}, the effect of \textit{contextual} embedding on the model performance  has been shown. \acrshort{ince} outperforms solutions that just use \textit{columnar} embedding and, more generally, produces results that are on par with or even better than those of \acrshort{sota} \acrshort{dl} models when applied to tabular data.

In this subsection, we visually examine how this mechanism improves the features representation, enhancing the performance of the final model. For sake of simplicity, in the following discussion, we use the Titanic \cite{dua2019uci} dataset. The supervised learning problem is a binary classification. The preprocessed dataset contains eight features. $\text{Age}$ and $\text{fare}$ are the zero-mean and one-standard-deviation continuous variables. The categorical features are $\text{sex} \in \left\lbrace \text{female},\ \text{male}\right\rbrace $, $\text{title} \in \left\lbrace \text{Mr.},\ \text{Mrs.},\ \text{Rare}\right\rbrace $, $\text{pclass} \in \left\lbrace 1,\ 2,\ 3\right\rbrace $, $\text{family\_ size} \in \left\lbrace 0,\ 1,\ 2,\ 3,\ 4,\ 6,\ 7,\ 8\right\rbrace $ , $\text{is\_alone} \in \left\lbrace 0,\ 1\right\rbrace $, $\text{embarked} \in \left\lbrace \text{C}=\text{Cherbourg},\ \text{Q}=\text{Queenstown},\ \text{S}=\text{Southampton}\right\rbrace $. For this exercise, we consider a simple \acrshort{ince} model with latent space size $l=2$, $\text{\acrshort{mlp}}_{\text{N, E}}$ depth $d=3$ and $n=2$. The choice of $l=2$ allows analyzing the representation in latent space without alleged artifacts introduced by the dimensional reduction. 

\begin{figure}[ht]
	\centering
	\includegraphics[width=\linewidth]{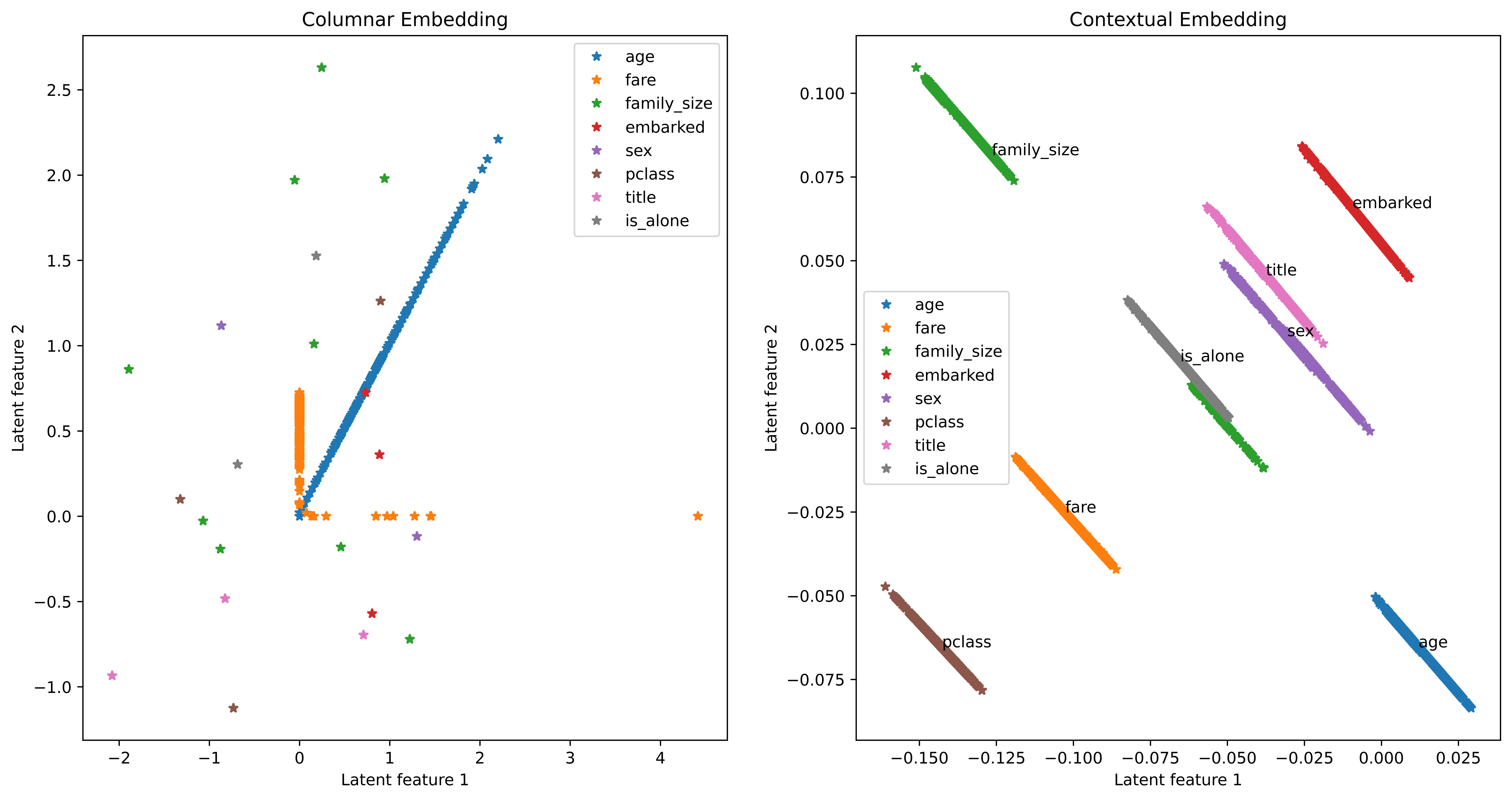}
	\caption{\textbf{Left:} \textit{Columnar} embedding before the stack of \acrshort{in}. \textbf{Right:} \textit{Contextual} embedding from the last \acrshort{in}.}
	\label{fig:columnar_vs_contextual_titanic}
\end{figure}

The left side plot of Fig. \ref{fig:columnar_vs_contextual_titanic} shows the output of \textit{columnar} embedding. Semantically related features like $\text{pclass}$-$\text{fare}$, $\text{sex}$-$\text{title}$, or $\text{family\_size}$-$\text{is\_alone}$ are distributed without any discernible pattern. The representation does not depend on the context: regardless of $\text{pclass}$, $\text{age}$ or $\text{family\_size}$ values, $\text{title}=\text{Mrs}$ is always projected to the same point in the latent space size.         

The \textit{contextual} embedding is depicted in the right side plot of Fig. \ref{fig:columnar_vs_contextual_titanic}. This represents the message sent from each node (i.e. tabular feature) to update the $\text{\acrshort{cls}}$ representation in the last \acrshort{in}. Patterns are easily discernible:  $\text{sex}$ vs. $\text{title}$, $\text{family\_size}$ vs. $\text{is\_alone}$ and the feature that is closest to $\text{pclass}$ is $\text{fare}$. Additionally, it is feasible to see that the latent projections of categorical features are not yet limited to a fixed number of points when the context is taken into consideration, as shown, for example, by $\text{title}$ embedding.

\subsection{Feature importance from Feature-Feature interaction}
The attention map for the $\text{\acrshort{cls}}$ virtual node may be used to assess the feature relevance  when the \textit{contextual} embedding is a Transformer \cite{gorishniy2021revisiting, somepalli2021saint}. Here, we look into if the feature-feature interaction that the \acrshort{in} learns can reveal details about the significance of tabular features. We first explain our methodology using the Titanic dataset for the purpose of simplicity, and then we illustrate the findings we achieved using the same technique on the other tabular datasets.     

In contrast to the Transformer case, we now have two new problems to resolve: 1) The feature-feature interaction is a $l$-dimensional vector (that means, it is not a scalar); 2) To assess the feature global significance, we must aggregate the feature-feature importance. The description of our process is provided below. 

\textit{First Step:} We split data in train/test datasets. We train the model and use the trained \acrshort{ince} on the test dataset to produce the feature-feature interaction, i.e. $e^{i}_{j_1 \rightarrow j_2} \in \mathbb{R}^l$ in Eq. \ref{eq:edge_update} returned by the last \acrshort{in}. In this notation, we have explicitly recovered the tabular row index $i$. 

\textit{Second Step:} We estimate mean \textbf{$\mu$} and covariance \textbf{$S$} of the entire population $\left\{e^{i}_{j_1 \rightarrow j_2}\right\}$ $\forall\, i, j_1, j_2$. 

\textit{Third Step:} For each pair $(j_1, j_2)$ of features and for each test row $i$, we compute the squared Mahalanobis distance:

\begin{equation}
D_i^2 \left(j_1\rightarrow j_2\right) = \left(e^{i}_{j_1 \rightarrow j_2} - \mu \right) S^{-1} \left( e^{i}_{j_1 \rightarrow j_2} - \mu \right) \nonumber \,.
\end{equation}

\textit{Fourth Step:} The squared Mahalanobis distance follows a Chi-Square distribution, so we can normalize the distance using p-value. The number of degrees of freedom of Chi-Square is given by the latent space size $l$:

\begin{equation}
p_i \left(j_1\rightarrow j_2\right) = \text{Pr} \left( D_i^2 \geq \chi_{l}^{2}  \right) \nonumber \,.
\end{equation}

\textit{Fifth Step:} The global interaction p-value $p \left(j_1\rightarrow j_2\right)$  is obtained averaging the previous results over the test dataset: 
\begin{equation}
p \left(j_1\rightarrow j_2\right) = \frac{1}{N_{test}} \sum_{i=1}^{N_{test}} p_i \left(j_1\rightarrow j_2\right) \nonumber \,.
\end{equation} 

\begin{figure}[ht]
	\centering
	\includegraphics[width=\linewidth]{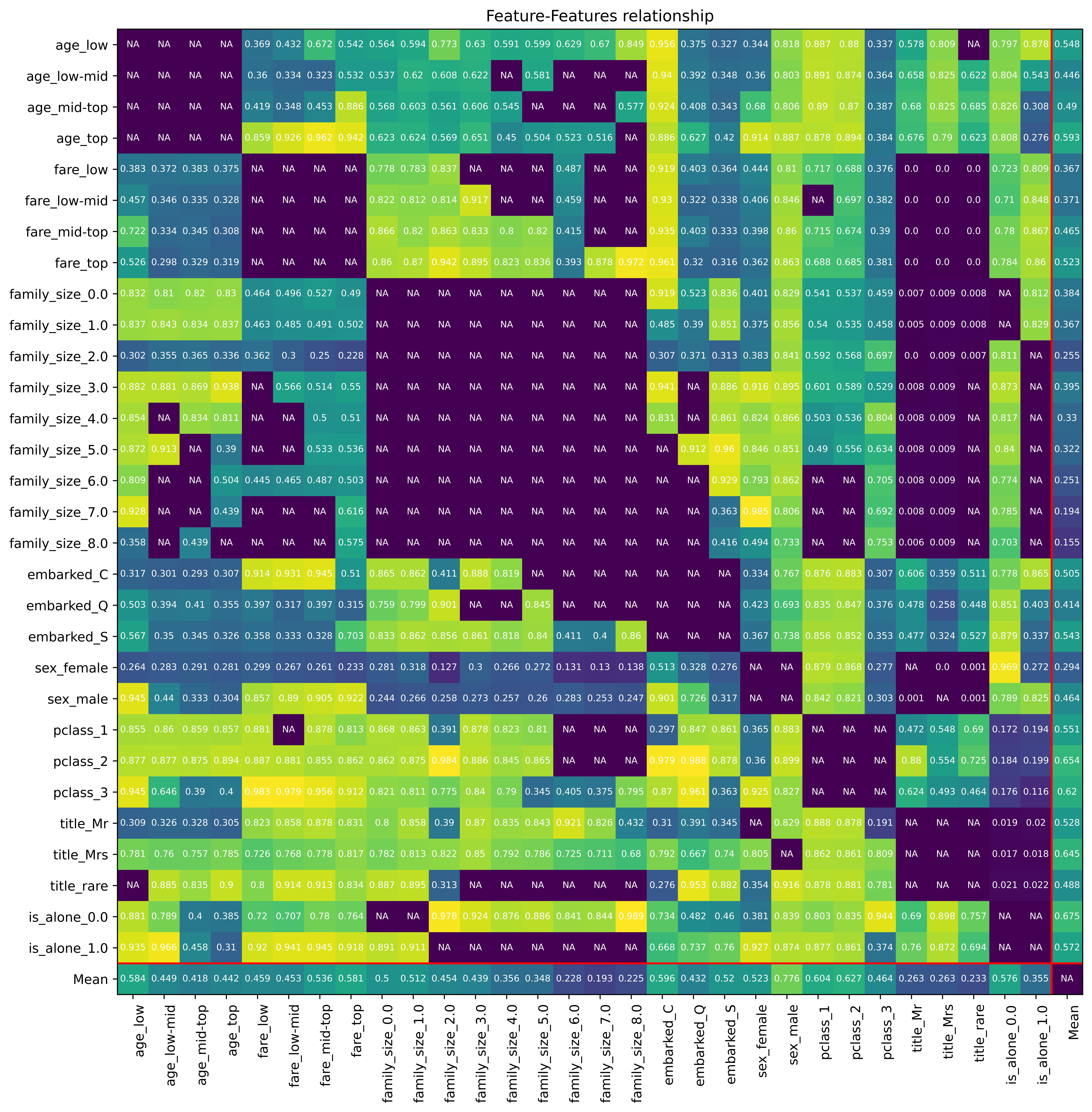}
	\caption{Titanic feature-feature interaction at feature-value level.}
	\label{fig:feature_val_feature_val_titanic}
\end{figure}

The findings of the proposed methodology on the Titanic dataset are displayed in the heatmap of Fig. \ref{fig:feature_val_feature_val_titanic}. The results are broken down at the feature-value level (i.e. $\text{sex}=\text{female}$, $\text{sex}=\text{male}$, $\text{title}=\text{Mrs}$, $\text{title}=\text{Mr}$, etc.). This is how the heatmap may be understood: the relevance of the message from the row-$r$-feature to the column-$c$-feature is represented by the element (row=$r$, column=$c$) of the heatmap. A lower p-value implies more significance. The last column, "Mean", is created by averaging all of the row values and shows the average relevance of the messages sent by row-$r$-feature. In a similar way, the last row (also known as "Mean") is derived by averaging all the values of the columns and it represents the mean relevance of the messages received by column-$c$-feature.

In order to quantitatively assess the quality of the heatmap, we compute the Spearman Rank correlation $\rho$ between 
\begin{equation}
p(j) = \frac{1}{|\mathcal{N}|}\sum_{\hat{j} \in \mathcal{N}} p(j, \hat{j}) = \frac{1}{2} \left[p \left(j\rightarrow \hat{j}\right) + p \left(\hat{j}\rightarrow j\right)\right] \,,
\nonumber
\end{equation}
and the feature importance calculated by KernelShap \cite{lundberg2017aunified}. In the formula above, $\mathcal{N}$ and $|\mathcal{N}|$ are the set of neighbors of node $j$ and its size, respectively. The outcome for the Titanic dataset is $\rho = 0.81 ( \text{p-value} = 0.05)$.

The heatmap and the Spearman Rank correlation provide the following insights. 

\textbf{a)} The feature-feature interaction is not symmetric. In the fully connected graph we have two independent edges $j_1\rightarrow j_2$ and $j_2\rightarrow j_1$ and the Eq. \ref{eq:mlp_e} is not invariant by $j_1 \longleftrightarrow j_2$ interchange. Our experiments demonstrate that inducing  $j_1 \longleftrightarrow j_2$ invariance in Eq. \ref{eq:mlp_e} results in a learning bias that negatively affects \acrshort{ince} performance. 

\textbf{b)} From heatmap, it is possible to discern logical patterns. For example, $\text{is\_alone}=1$ does not add information (high p-value) when $\text{family\_size}$ is $0$ or $1$ and on the contrary, the value $\text{family\_size}$ is very relevant (low p-value) for any value of $\text{title}$.  

\textbf{c)} Considering that KernelShap evaluates global model behavior (including the decoder) and that \acrshort{in} models separately $\left(j_1 \rightarrow j_2 \right)$ and $\left(j_2 \rightarrow j_1 \right)$ and that we have to aggregate and average them to compare with KernelShap, the Spearman Rank correlation analysis result can be considered encouraging.

Finally, Table \ref{tab:spearman} summarizes the Spearman Rank correlation achieved on various datasets and demonstrates how the results are consistent regardless of the dataset under consideration.

\begin{table*}
	\caption{Spearman Rank Correlation between KernelShap and feature-feature interaction}
	\label{tab:spearman}
	\begin{center}
		\begin{tabular}{ccccc}
			\toprule
			& \acrshort{heloc} & Cal. Hous. & Adult Inc. & Forest Cov. \\
			\midrule
			$\rho (\text{p-value})$ & $0.82 (0.04)$  & $0.80 (0.06)$ & $0.85 (0.03)$ & $0.81 (0.04)$\\
			\bottomrule
		\end{tabular}
	\end{center}
\end{table*}

\section{Conclusions}
\label{sec:conclusions}
Let us highlight the main contributions of this article:
\begin{itemize}
	\item As far as we know, this is the first time that model architecture proposes the use of \acrshort{gnn} for contextual embedding to solve supervised tasks involving tabular data.
	\item Literature discusses mainly about the usage of Transformers. This manuscript shows that \acrshort{gnn}, particularly \acrshort{in}, are a valid alternative. It shows better performance with a lower number of training parameters.
	\item As a matter of fact, this innovative architecture outperforms the state of the art \acrshort{dl} benchmark based on 5 different diverse datasets. Moreover, it closes the gap with classical \acrshort{ml} models (tree-based), outperforming them in 2 of these datasets, and being very close in two more. The tradeoff versus tree-based models is additional computational load in the form of training time, and scalability issues with the number of features (nodes) of the dataset, which constitute future lines of research to keep improving its practical implementation.
	\item Finally, the interpretability of \acrshort{gnn} is explored. This is a key topic for industry environments, and apparently this is the first study for \acrshort{gnn} and tabular data.
\end{itemize}

\appendix
\section{Normalized Metric}
\label{sec:normalization_metric}

\begin{algorithm}
	\caption{Normalized metric}\label{alg:cap}
	\begin{algorithmic}
		\renewcommand{\algorithmicrequire}{\textbf{Input:}}
		\renewcommand{\algorithmicensure}{\textbf{Output:}}
		\REQUIRE $l\colon$ latent space, $r\colon$ dataset
		\ENSURE $C_d$, $C_n$ two list of normalized metric
		\STATE $base \gets Metric_{r} ( d=1, n=1, l, r)$
		\STATE $C_d \gets Metric_{r} ( d, n=1, l, r)$ \ \ \ \ \ $\forall d \in \left\lbrace 1,2,3,4\right\rbrace$
		\STATE $C_n \gets Metric_{r} ( d=1, n, l, r)$ \ \ \ \ \ $\forall n \in \left\lbrace 1,2,3,4\right\rbrace$
		\STATE $best \gets Best_{r} ( C_d, C_n)$
		\STATE $C_d \gets \dfrac{C_d - base}{best - base}$ \ \ \ \ \ $\forall d \in \left\lbrace 1,2,3,4\right\rbrace$
		\STATE $C_n \gets \dfrac{C_n - base}{best - base}$ \ \ \ \ \ $\forall n \in \left\lbrace 1,2,3,4\right\rbrace$
		\RETURN $C_d$, $C_n$
	\end{algorithmic}
\end{algorithm}
\noindent $\forall l,r /$
$r \in \left\lbrace \right.$\acrshort{heloc}, Cal. Hous., Adult Inc., Forest Cov., HIGGS$\left.\right\rbrace$, $l \in \left\lbrace 16, 32, 64, 128\right\rbrace$. In Alg. \ref{alg:cap} $Metric_{r}$ and $Best_r$ are Accuracy/\acrshort{mse} and $max$/$min$ depending on $r$, $d$ is the $\text{\acrshort{mlp}}_{\text{N, E}}$ depth  and $n$ is the number of stacked \acrshort{in}. Notice that computing $Metric_{r}$ means train-test the model 5 times with different seeds and average the results.

The curves of Fig. \ref{fig:performance_analysis} are obtained by computing the average and the standard-deviation from results of Alg. \ref{alg:cap}.



%%       \citet{<label>} ==> Jones et al. [21]
%%       \cite{<label>} ==> [21]
%%

\bibliographystyle{elsarticle-num-names.bst} 
\bibliography{bibliography}


%\begin{thebibliography}{00}



%\end{thebibliography}
\end{document}

%% file: glossaries.tex
\newacronym{ai}{AI}{Artificial Intelligence}
\newacronym{dl}{DL}{Deep Learning}
\newacronym{ml}{ML}{Machine Learning}
\newacronym{gbdt}{GBDT}{Gradient Boosting Decision Tree}
\newacronym{gnn}{GNN}{Graph Neural Network}
\newacronym{ignn}{IGNN}{Interaction Graph Neural Network}
\newacronym{xgboost}{XGBoost}{eXtreme Gradient Boosting}
\newacronym{lightgbm}{LightGBM}{Light Gradient Boosting Machine}
\newacronym{heloc}{HELOC}{Home Equity Line of Credit}
\newacronym{saint}{SAINT}{Self-Attention and Intersample Attention Transformer}
\newacronym{mse}{MSE}{Mean Squared Error}
\newacronym{catboost}{CatBoost}{Categorical Boosting}
\newacronym{ince}{INCE}{\acrlong{in} Contextual Embedding}
\newacronym{mlp}{MLP}{Multi-Layer Perceptron}
\newacronym{in}{IN}{Interaction Network}
\newacronym{sota}{SOTA}{state of the art}
\newacronym{cls}{CLS}{CLS}
\newacronym{bert}{BERT}{Bidirectional Encoder Representations from Transformers}
\newacronym{node}{NODE}{Neural Oblivious Decision Ensembles}
\newacronym{fttransformer}{FT-Transformer}{Feature Tokenizer + Transformer}
\newacronym{netdnf}{Net-DNF}{Disjunctive Normal Form Network}